# Finetuning Randomized Heuristic Search For 2D Path Planning: Finding The Best Input Parameters For R* Algorithm Through Series Of Experiments

Konstantin Yakovlev, Egor Baskin, Ivan Hramoin

Institute for Systems Analysis of Russian Academy of Sciences, Moscow, Russia
{yakovlev, baskin, hramoin}@isa.ru

**Abstract.** Path planning is typically considered in Artificial Intelligence as a graph searching problem and R* is state-of-the-art algorithm tailored to solve it. The algorithm decomposes given path finding task into the series of subtasks each of which can be easily (in computational sense) solved by well-known methods (such as A*). Parameterized random choice is used to perform the decomposition and as a result R* performance largely depends on the choice of its input parameters. In our work we formulate a range of assumptions concerning possible upper and lower bounds of R* parameters, their interdependency and their influence on R* performance. Then we evaluate these assumptions by running a large number of experiments. As a result we formulate a set of heuristic rules which can be used to initialize the values of R* parameters in a way that leads to algorithm's best performance.

## 1    Introduction

Ability to plan a path is one of the key features for an intelligent agent. In our work, we examine the case when an agent operates in a rectangle-bounded region of static 2D environment composed of traversable and non-traversable areas (free space and obstacles). We use 8-connected grid as a formal model of agent's environment [1, 2, 3]. Within this model the task is to find a sequence of unoccupied adjacent grid cells connecting given start and goal cells.

Heuristic search algorithm A* [4] is widely used in AI community for finding paths on grids. A* using admissible heuristics guarantees finding a shortest path [5]. Plenty of such heuristics for grid-worlds exist: Manhattan distance, octile (diagonal) distance etc. These heuristics being natural metrics in grid-worlds are the "best-available", but nevertheless A*-search exercising them explores too much of the state-space in case the goal is located beyond the obstacles. The reason of the over-exploration is that A* guided locally by one of the abovementioned heuristics necessarily (in presence of obstacles) falls into a local minimum – "the portion of state-space from which there is no way to states with smaller heuristics without passing through states with higher heuristics" [6]. There exist a number of approaches to reduce the A* search space (and thus increase the computational effectiveness of path planning algorithm). One approach is to modify A* in some way. Using weighted heuristics [7, 8, 9], implementing iterative deepening techniques [10, 11], imposing limits on the size of the set of candidate cells for exploration [12, 13] are examples of such approach. Another approach exploits the idea of decomposition. Methods implementing this approach split the given task to the series of subtasks (local tasks) each of which is solved independently (by local planners) and final solution is constructed by the composition of local solutions. Decomposition can be performed using predefined criteria [14] or in random fashion [15, 16].

One of the well-known state-of-the art algorithms suitable for path finding in grid-worlds based on parameterized random decomposition is R* introduced in [6]. It exploits the same ideas lying behind RRT planners [16] and uses WA* (A* with weighted heuristic) as local planner. To perform a search R* needs to be provided with the values of its 3 input parameters (as well as the weight of the heuristic used for the local WA* search). Preliminary experiments show that the algorithm performance largely depends on these values. At the same time to the best of authors knowledge there is no reported research results on how exactly the choice of the parameters values influence the performance of R* and which values should be used to solve practical path planning tasks with R*. This works aims at filling this gap.

In our work we theoretically analyze the possible influence of each parameter on R* performance along with evaluating its lower and upper bounds. We show that the bounds for 2 parameters are either constants or can be expressed as functions of the 3rd parameter. At the same time, we propose that the value of the latter is the function of start and goal positions. Then we perform comprehensive experimental analysis of R* solving more than 5000 of 2D path planning tasks to estimate the coefficients of the parameters' bindings proposed before. Thus we end up with a set of rules for R* parameterization applying which leads to algorithm's best performance.

## 2 R* algorithm for grid path planning

### 2.1 Path planning problem

Consider a 8-connected grid which is a finite set of cells $A=(a, b, c, …)$ that can be represented as a matrix $A_{M \times N}=\{a_{ij}\}$, where: $i, j$ – are cell position indexes (also denoted as $i(a), j(a)$) and $M, N$ – are grid dimensions. Each cell is labeled either traversable or un-traversable and agent is allowed to move from a traversable cell to one of its traversable neighbors.

A metric function *dist* (also known to AI community as diagonal heuristic) is used to measure the distance between any two cells:
$$dist(a_{ij}, a_{kl}) = c_d \cdot \min(\Delta_i, \Delta_j) + c_{hv} \cdot (\Delta_i + \Delta_j - 2 \cdot \min(\Delta_i, \Delta_j)),$$
where $\Delta_i=|i–k|$, $\Delta_j=|j–l|$, $c_d=k \cdot c_{hv}$ ($1<k<2$), $c_{hv}=const \in \mathbf{R}^+$. $c_{hv}$ is a distance between a cell and any of its horizontal or vertical neighbors, $c_d$ is a distance between a cell and any of its diagonal neighbors. In our work we use integer constants $c_{hv}=10$ and $c_d=14$ for corresponding distances.

Path planning task is considered to be set if two distinct traversable cells – start and goal – $s, g \in A$ are set. The solution of the problem is a path $\pi(s, g)$, *e.g.* a sequence of traversable adjacent cells starting with $s$ and ending with $g$. The length of the path $L(\pi)$ is the sum of the distances between all pairs of adjacent cells forming the path.

## 2.2 Randomized Heuristic Search Algorithm R* Overview

R* – is state-of-the art heuristic search algorithm that decomposes initial path planning task into series of subtasks, identifies ones to be solved by local planner (WA*) and tries to solve them. If the local solution is found "easily" it is stored and can be used lately to reconstruct final solution but if the local solution is "hard to find" R* postpones local search and chooses another local task. As its creators say in [6]: "R* postpones the ones [*local searches*] that do not find solutions easily and tries to construct the overall solution using only the results of searches that find solutions easily".

We encourage the reader to examine the papers [6, 17] for the detailed explanation of R* and now give only a brief overview of the algorithm.

At each step R* firstly chooses the most promising cell $c$ from OPEN list (initially containing only start cell). Then algorithm randomly selects $K$ traversable cells residing at the distance $\Delta$ from $c$ and inserts them into OPEN. These cells ($b_i$) are called the successors of $c$, while $c$ itself is called the predecessor ($pred(b_i)=c$). If $dist(g, c) \leq \Delta$ then the goal cell is also added to OPEN.

Next R* tries to find a local path $\pi(pred(c), c)$ with WA* algorithm. If the path is not found after the *m* steps of WA* the cell $c$ is labeled AVOID which means it was hard to find the current local path and the local search should be postponed. Cell $c$ is kept in OPEN list in that case. If the path is found the cell is removed from OPEN and is inserted into CLOSED list.

The process of generating successors, adding them to OPEN, choosing the best cell in OPEN and trying to find a local path is referred as the expansion of the cell $c$.

R* chooses the cells from OPEN using the same heuristic rule as WA*. The only difference is that R* chooses only such a cell which is not labeled AVOID (initially none of the cells has this label) and only if no such cells are left in OPEN it chooses a cell amongst AVOID ones. The stop criteria for R* is analogous to WA*.

One specific thing related to algorithm's implementation which is not addressed in original paper is the procedure of generating successors for the expanding cell. In our implementation of R* we use midpoint circle algorithm [18] to generate successors – see fig.1.

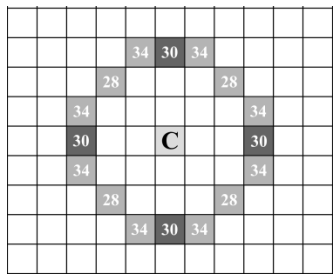

**Fig. 1.** Set of successors for cell *C*. Numbers indicate the distance from *C*.

One can think that the circumference of radius $\Delta/c_{hv}$ (if measured in cells) with the center in the cell $c$ is "drawn" and $K$ traversable cells forming this circumference are randomly chosen as successors of $c$.

## 2.3 R* Parameters Influence on Algorithm Performance and Their Lower and Upper Bounds

Obviously R* performance depends on the values of its 3 parameters: $K$ – number of successors generated for each expanded cell $c$, $\Delta$ – the distance between $c$ and generated successors, $m$ – number of steps local planner is allowed to perform before abandoning the search. Let's analyze the influence of each parameter and assess its lower and upper bounds taking into account R* working principles.

The value of $m$ affects both execution time and memory usage. The higher the value of $m$ is the more steps are performed by WA* while finding *each* local path. At the same time it is likely that only small fraction of all local paths compound final solution which means in case of higher $m$ more chances are that R* wastes much time on "useless" computations. Thus high values of $m$ should be avoided.

The influence of parameter $m$ on path length is less evident. It is likely that the latter depends primarily on how local goal cells are chosen at each step of R* and that is independent of $m$.

The lower bound for $m$ can be assessed in the following way. One can show that minimum number of steps WA* needs to find a path is $m^*=\max(|i(s)-i(g)|, |j(s)-j(g)|)$. This happens in particular when there are no obstacles in between $s$ and $g$. As detailed above our implementation of R* uses midpoint circle algorithm to form the set of possible successors and it can be shown that $\max(|i(s)-i(g)|, |j(s)-j(g)|)$ is achieved when $s$ (the cell under expansion) and $g$ (the successor of $s$) lie on the same grid row (or column) and in that case $m^*=\Delta/c_{hv}$. So the value $m'=\Delta/c_{hv}(=\Delta/10)$ is the lower bound for $m$ (if $m<\Delta/10$ WA* would simply fail to solve local path finding tasks in most cases).

Theoretically upper bound for $m$ is the number of all traversable cells on grid but it's reasonable to limit $m$ by some value $m''=k \cdot m'$ ($k \in N$) which means that the local planner (WA*) is allowed to perform $k$ times more steps to solve local path planning tasks than in the most trivial cases (when there are no obstacles present).

The value of $K$ affects both execution time and memory usage but primarily memory usage as *all* $K$ successors for *each* expanded cell are permanently stored in memory. So, if we are interested in decreasing memory consumption high values of $K$ should be avoided. At the same time, the higher the value of $K$ is the more successors for each cell are generated and more chances are R* would pick up "good" candidates for further expansion (the candidates that minimize both local and overall path lengths). So setting $K$ to high values potentially leads to better quality solutions.

Upper bound for $K$ can be assessed in the following way. As said above, set of successors for any expanded grid cell is the subset of cells comprising the discrete circumference of radius $\Delta/c_{hv}$ (if measured in cells). The length of such circumference equals $2 \cdot \pi \cdot \Delta/c_{hv}$. Thus maximum number of successors $K'' \approx 6 \cdot \Delta/10$. Minimum possible number of successors $K'$ apparently equals 1. But it's obvious that in that case we would likely get very awkward shaped and very long paths, so we suggest $K'=3$.

As shown above upper/lower bounds for $m$ and $K$ can be expressed in relation to $\Delta$ making $\Delta$ the key parameter for R* which value affects both execution time and memory usage and solution quality. Considering the influence of $\Delta$ independently one can say that the higher the value of **Δ** is (with *dist*(*s*, *g*) being the maximum) the more R* relies on local path planning which makes algorithm behave more like typical A*-family algorithm (the behavior we are trying to avoid). On the other hand setting **Δ** to extremely small values (with $c_{hv}$=10 being the minimum) directly converges R* to A* which does not make any sense at all. So, the value of **Δ** should be picked from the middle of the spectrum of possible values. In other words, **Δ** can be represented as a positive monotone function (with known minimum and maximum) of start and goal locations, e.g. **Δ**=*dist*(*s*, *g*)/*k* and the "right" value for binding coefficient *k* should be estimated via experimental analysis but it is likely to belong to the middle range of possible spectrum (defined by minimum and maximum values of **Δ**).

By now we have theoretically evaluated lower and upper bounds for the R* parameters (***m***, **K** and **Δ**) and showed that ***m*** and **K** can be expressed as linear functions of **Δ** and **Δ** can be expressed as linear function of start and goal locations. The coefficients of the bindings are unknown and we are going to evaluate them experimentally.

## 3      Experimental analysis

### 3.1    Testbed

To examine the influence of R* input parameters on the algorithm's performance and find parameters "best" values we have run 5250 of experiments on 3 types of grids:
- randomly generated grids containing rectangle shaped obstacles of different sizes (70 grids * 25 different parameters configurations = 1750 experiments);
- randomly generated grids containing tetris-shaped obstacles of different sizes (70 grids * 25 different parameters configurations = 1750 experiments);
- grids which are models of city landscape (70 grids * 25 different parameters configurations = 1750 experiments).

While generating grids containing rectangle and tetris-shaped obstacles the latter were added one by one at randomly selected positions until the total number of untraversble cells equals or slightly exceeds predefined threshold, *e.g.* 30%. When adding each obstacle its size and orientation was chosen randomly within predefined thresholds.

Grids modeling city landscape were generated semi-automatically and the maps of the real cities were used as sources. The percentage of blocked cells on these grids equals or slightly exceeds 30% (just as on randomly generated grids).

All grids were of the size 501x501 and start and goal cells were always located on the opposite edges of grid in such a way that *dist*(*s*, *g*)=5000.

The following indicators were used to evaluate R* performance:
*cells* – the number of cells stored in OPEN and CLOSED (used to assess memory consumption);
*time* – time (in ms) used by R* to find a path;
*length* – length of the path found.

### 3.2 Results

There were conducted 3 consecutive series of experiments and a preliminary one.

Preliminary experiments were aimed at fine tuning local planer (WA*), *e.g.* estimating the best value for weight of heuristic function. For the sake of space we omit the results of experiments but they count in favor value 3 should be used as the weight. As R* is supposed to use the same heuristic function as local planner weighted by the same weight (to guarantee suboptimality) the later was also set to 3.

First, we examined the influence of parameter *m* – number of steps local planner (WA*) is allowed to commit before abandoning the search for a local path – on R* performance. Then parameter *K* – number of successors generated for each expanded cell – was evaluated with *m* being set to its best value. Finally, parameter $\Delta$ – the distance between expanded cell and its successors – was evaluated (with *m* and *K* being fixed to their best values discovered before).

The averaged results of the experiments are shown on fig. 2. These averaged results correlate well with all "individual" ones, *e.g.* results obtained on grids of specific types, and thus can be used as consistent basis for R* performance evaluation.

In the first series of experiments values of $\Delta$ and *K* were set to *dist*(*s*, *g*)/10(=500) and $\Delta$/10(=50) respectively and *m* was assigned a range of values: 50, 75, 100, 200, 300, 500, 750, 1000.

Obtained results (see fig. 2) support the assumption that *m* mainly affects running time and memory consumption – *time* and *cells* values differ (due to different *m*-values) ≈5 and 1,5 times respectively – while the influence of *m* on solution quality is less evident. Interesting case which breaks the evident tendency – the higher the value of *m* is the worse the performance of R* is – is setting *m* to 50. Worst results in that case (which are not depicted on diagrams but reported in the table) can be easily explained: when *m* is set to 50 local planner is guaranteed to almost always fail in local pathfinding (during the first attempt). So in the end more such searches are performed which in turn substantially degrades R* performance.

Based on gained results value 100(=$\Delta$/5) can be recommended to initialize *m*. This can be interpreted in the following way (see previous section): minimum number of steps local planer needs to find a path is $\Delta$/10, so to get "best" results local planner should be allowed to use 2 times more steps (than the minimum).

In the second series of experiments value of $\Delta$ remained the same, *m* was set to its best value, *e.g* $\Delta$/5, and *K* was assigned a range of values: 3, 5, 7, 10, 25, 50, 70, 100.

Obtained results support the assumption that *K* drastically affects memory consumption and running time – *cells* and *time* values differ 4 and 3 times (due to different *K*-values) respectively; they also justify that *K* has a major influence on solution quality– difference in *length* reaches 30-35%.

Results of the experiments evidently show that higher values of *K* should be avoided due to high computational costs and lower values of *K* should be avoided due to lower solution quality. On this basis the recommended value for *K* is 25=($\Delta$/20). This can be interpreted as following: maximum number of successors for any cell ≈6·$\Delta$/10, so to get best results R* should generate round $1/10^{th}$ -$1/12^{th}$ of that value (1 out the 10-12 possible successors should be generated).

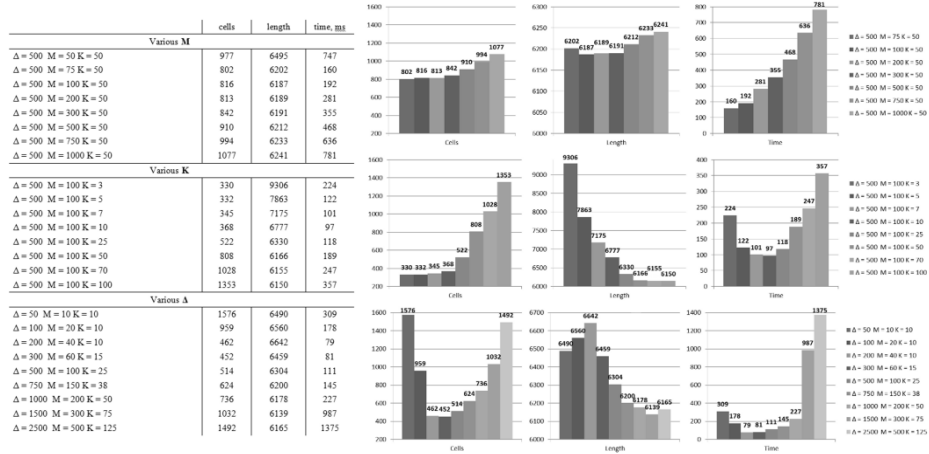

**Fig. 2.** Experimental results.

In the third series of experiments *m* and *K* were assigned their best values, *e.g* $\Delta/5$ and $\Delta/20$, and $\Delta$ was consequently initialized as: 50, 100, 200, 300, 500, 750, 1000, 1500, 2500.

Obtained results (see fig. 2) verify the assumption that $\Delta$ (just like *K*) has more influence on R* performance rather than *m*: *cells* and *time* differ 4 and 10 times respectively (due to the different values of $\Delta$) and the difference in *length* reaches 10%.

As one can see, setting $\Delta$ to higher or lower values significantly reduces algorithm's computational efficiency (the fact we have predicted earlier) so these values should be rejected and the values from the middle of the spectrum should be used. Based on the obtained results we recommend setting $\Delta$ to 500(=*dist*(*s*, *g*)/10).

Summarizing the results of experimental analysis two main conclusions can be made. First, R* performance depends *largely* on the values of its parameters and assigning them in a wrong way can lead to a dramatic fall in computational efficiency. Second, there exist a set of rules which can be used to automatically initialize R* input parameters in such a way that leads to best performance. These rules can be formalized as the set of bindings:

$$\Delta = dist(s, g)/10;$$
$$K = \max(10, \Delta/20);$$
$$m = \Delta/5.$$

Presented bindings are dependent only on start and goal locations which are known a priori and thus can be viewed as a universal (heuristic) rule of parameterizing R* when solving path planning tasks on 8-connected grids.

## 4 Conclusions

R* is state-of-the-art randomized heuristic search algorithm and a powerful tool to solve 2D path planning tasks at low computational costs. But to benefit from using R* in actual practice one needs to initialize 3 algorithm's parameters in a "right" way as

R* performance heavily depends on them. In presented work we analyze (both theoretically and experimentally) the nature of that dependencies and end up with the set of heuristic rules that can be used to automatically parameterize R* in order to get the best results (low computational cost and high solution quality). Presented rules are easily applicable to any path planning task in any grid-world as they do not require any additional knowledge except the positions of start and goal cells.

**References**
1. Elfes, A. 1989. Using occupancy grids for mobile robot perception and navigation. Computer, 22(6), 46-57.
2. Yap, P. 2002. Grid-based path-finding. In Proceedings of 15th Conference of the Canadian Society for Computational Studies of Intelligence, 44-55. Springer Berlin Heidelberg.
3. Tozour, P. 2004. Search space representations. In Rabin, S. (Ed.), AI Game Programming Wisdom 2, 85–102. Charles River Media.
4. Hart, P. E., Nilsson, N. J., & Raphael, B. 1968. A formal basis for the heuristic determination of minimum cost paths. IEEE Transactions on Systems Science and Cybernetics, 4(2), 100-107.
5. Pearl, J. 1984. Heuristics: intelligent search strategies for computer problem solving. Addison-Wesley.
6. Likhachev, M., & Stentz, A. 2008, R* Search. In Proceedings of the Twenty-Third AAAI Conference on Artificial Intelligence. Menlo Park, Calif.: AAAI press.
7. Pohl, I. 1970. First results on the effect of error in heuristic search. In Bernard M., Michie D. (Eds.), Machine Intelligence 5, 219–236. Edinburg: Edinburgh University Press.
8. Gallab, M., & Dennis A. 1983. A$\varepsilon$ – an efficient near admissible heuristic search algorithm. In Proceedings of the Eighth International Joint Conference on Artificial Intelligence (IJCAI-83), 789–791.
9. Likhachev M., Gordon G., & Thrun S. 2004. ARA*: Anytime A* with Provable Bounds on Sub-Optimality, Advances in Neural Information Processing Systems 16 (NIPS). Cambridge, MA: MIT Press.
10. Korf, R. E. 1985. Depth-first iterative-deepening: An optimal admissible tree search. Artificial intelligence, 27(1), 97-109.
11. Reinefeld, A., & Marsland, T. A. 1994. Enhanced iterative-deepening search. IEEE Transactions on Pattern Analysis and Machine Intelligence, 16(7), 701-710.
12. Bisiani, R. 1987. Beam search. In Shapiro, S. (ed.), Encyclopedia of Artificial Intelligence, 56-58. John Wiley and Sons.
13. Zhang, W. 1998. Complete anytime beam search. In Proceedings of the Fifteenth National Conference on Artificial Intelligence (AAAI-98), 425–430.
14. Botea, A., Muller, M., & Schaeffer, J. 2004. Near optimal hierarchical path finding. Journal of game development, 1(1), 7-28.
15. Kavraki, L. E., Svestka, P., Latombe, J. C., & Overmars, M. H. 1996. Probabilistic roadmaps for path planning in high-dimensional configuration spaces. IEEE Transactions on Robotics and Automation, 12(4), 566-580.
16. LaValle S.M., 1998. Rapidly-exploring random trees: A new tool for path planning, Technical Report, 98-11, Computer Science Dept., Iowa State University.
17. Likhachev, M., & Stentz, A. 2008b. R* search: The proofs. Technical Report, University of Pennsylvania, Philadelphia, PA.
18. Pitteway, M. L. V. 1985. Algorithms of conic generation. In Fundamental algorithms for computer graphics, 219-237. Springer Berlin Heidelberg.